\documentclass{ecai} 

\usepackage{algorithm}
\usepackage{algorithmic}
\usepackage{multirow}
\usepackage{latexsym}
\usepackage{amssymb}
\usepackage{amsmath}
\usepackage{amsthm}
\usepackage{booktabs}
\usepackage{enumitem}
\usepackage{graphicx}
\usepackage{color}

\usepackage{soul, color, xcolor}

\newcommand{\BibTeX}{B\kern-.05em{\sc i\kern-.025em b}\kern-.08em\TeX}

\begin{document}


\begin{frontmatter}


\paperid{2202} 


\title{Robust Deep Hawkes Process under Label Noise of Both Event and Occurrence}


\author[A]{\fnms{Xiaoyu}~\snm{Tan}\footnote{Equal contribution.}}
\author[B]{\fnms{Bin}~\snm{Li}\footnotemark}
\author[B]{\fnms{Xihe}~\snm{Qiu}\thanks{Corresponding Author. Email: qiuxihe1993@gmail.com}\footnotemark} 
\author[C]{\fnms{Jingjing}~\snm{Huang}} 
\author[D]{\fnms{Yinghui}~\snm{Xu}} 
\author[A]{\fnms{Wei}~\snm{Chu}} 

\address[A]{INF Technology (Shanghai) Co., Ltd.}
\address[B]{Shanghai University of Engineering Science}
\address[C]{ENT Institute and Department of Otorhinolaryngology, Eye \& ENT Hospital, Fudan University}
\address[D]{Fudan University}




\begin{abstract}
Integrating deep neural networks with the Hawkes process has significantly improved predictive capabilities in finance, health informatics, and information technology. Nevertheless, these models often face challenges in real-world settings, particularly due to substantial label noise. This issue is of significant concern in the medical field, where label noise can arise from delayed updates in electronic medical records or misdiagnoses, leading to increased prediction risks. Our research indicates that deep Hawkes process models exhibit reduced robustness when dealing with label noise, particularly when it affects both event types and timing. To address these challenges, we first investigate the influence of label noise in approximated intensity functions and present a novel framework, the \textbf{R}obust \textbf{D}eep \textbf{H}awkes \textbf{P}rocess (\textbf{RDHP}), to overcome the impact of label noise on the intensity function of Hawkes models, considering both the events and their occurrences. We tested RDHP using multiple open-source benchmarks with synthetic noise and conducted a case study on obstructive sleep apnea-hypopnea syndrome (OSAHS) in a real-world setting with inherent label noise. The results demonstrate that RDHP can effectively perform classification and regression tasks, even in the presence of noise related to events and their timing. To the best of our knowledge, this is the first study to successfully address both event and time label noise in deep Hawkes process models, offering a promising solution for medical applications, specifically in diagnosing OSAHS.
\end{abstract}

\end{frontmatter}


\section{Introduction}

Temporal point processes (TPP) are designed to predict the nature and timing of forthcoming events by understanding the patterns of their occurrence \cite{gonzalez2016spatio}. These models are particularly suited for scenarios where events happen asynchronously (at unpredictable intervals), are multimodal (involving diverse event types), and are influenced by past occurrences. As a distinctive TPP approach, the Hawkes process is designed to simulate a self-exciting process over time. Essentially, as one event happens, it increases the likelihood of subsequent events, though this increased intensity fades as time passes. The model's applications span high-frequency trading \cite{chen2013inference}, bioinformatics gene mapping \cite{petrovic2022machine}, earthquake forecasting \cite{clements2011residual}, traffic pattern analysis \cite{zhu2021spatio}, and predicting patient disease trajectories and timings \cite{ru2022sparse,Qiu_2023_ICCV}.

To estimate the likelihood of an event's occurrence, the Hawkes process employs an intensity function influenced by past events. Traditional Hawkes models, however, often presume that prior events invariably have a positive impact on upcoming ones. This presumption can limit the model's adaptability, sometimes yielding suboptimal outcomes with limited versatility \cite{zhang2020self,zuo2020transformer}. To address this, several approaches incorporating deep neural structures have been proposed to enhance the dynamic modeling and demonstrate improved results in various real-world data experiments \cite{zhang2020self,zuo2020transformer,jiang2021eduhawkes}. Yet, even with the benefits of deep neural networks, these models overlook the significant label noise present in real-world scenarios. In real-world settings, data often suffers from label noise, which can hinder the efficiency and reliability of the learning process. Such noise might arise from human labeling mistakes or intrinsic flaws in electronic logging systems \cite{frenay2013classification,lukasik2020does,song2022labelnoise}. Especially in the medical domain, label noise is primarily induced by the inherent subjectivity in diagnoses, expert errors in manual labeling, equipment variability, and the complexity of medical conditions \cite{frenay2014comprehensive}, which introduce significant risk in medical diagnosis predictions using the Hawkes process. 

To explore the impact of label noise on  both event types and timing, we first conduct an ablation study focusing on the errors in learning intensity functions. We compare datasets with clean, accurate labels against those with noisy labels to understand this influence. Our analysis in Section \ref{section:label} reveals that label noise, whether in events or timing, significantly impairs the learning of intensity functions, leading to errors in model predictions. Furthermore, when both types of label noise are present, the resultant error in learning intensity functions surpasses the sum of errors caused by each type of noise independently. This indicates a compounded effect, where the interaction between event and time noise amplifies the learning challenges more than if each noise type acted alone. The compounded impact of event and time noise not only challenges conventional label noise mitigation strategies but also emphasizes the importance of developing a more sophisticated approach to improve robustness under both label noise and influence.

To separately mitigate the influence of label noise from events and timing, reduce the compounded effects, and improve the robustness of the Hawkes process under the deep learning framework, we therefore propose a novel framework called Robust Deep Hawkes Process (\textbf{RDHP}) to improve the performance under label noise, especially in real-world medical scenarios. We conduct experiments on synthetic open-source datasets and a real medical dataset on obstructive sleep apnea-hypopnea syndrome (OSAHS), along with different sequence lengths and varied sequence types. We demonstrate that the state-of-the-art Hawkes process methods under a deep learning framework are insufficiently robust against under label noises. In contrast, the RDHP model effectively mitigates the perturbing effects of noise during the learning phase, consistently demonstrating superior performance in numerous benchmarks.

\begin{figure}[h]
    \centering
    \includegraphics[width=0.5\textwidth]{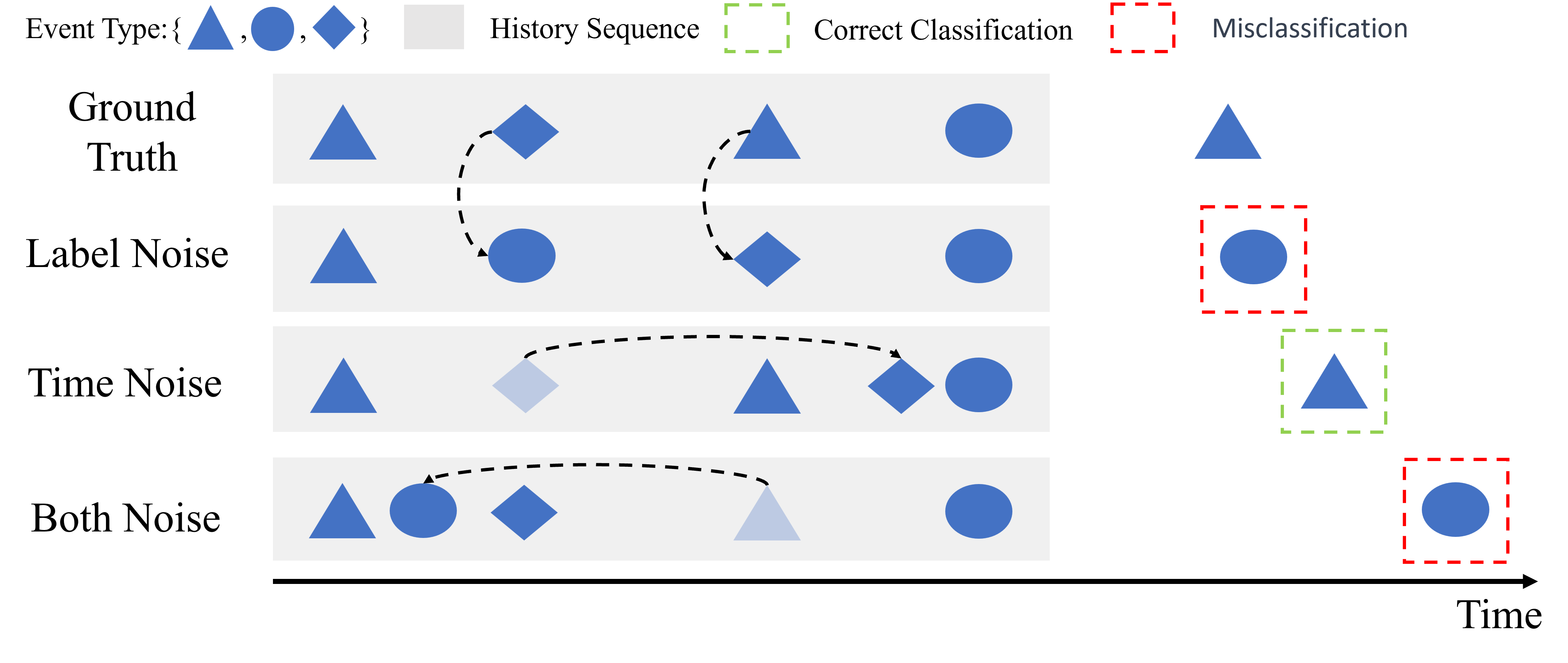}
        \caption{Label noise of both event and occurrence time points. Different shapes represent different types of events, and the noise affects the type of event and the occurrence time. The dashed lines represent the influence of noise.}
    \label{fig:noise}
\end{figure}

\vspace{15px}

The main contributions of our work can be summarized as follows:
\begin{itemize}
     \item 
     Current deep Hawkes models overlook the challenge of label noise. We investigate the effect of label noise from both events and occurrences and introduce the \textbf{RDHP}\footnote{The code is available at https://github.com/testlbin/RDHP} framework to effectively mitigate such an effect. To our best knowledge, this is among the first efforts to address this particular issue in deep Hawkes models.
     \item 
     We employ the generalized cross-entropy (GCE) loss function, sparse overparameterization, and other re-weighting techniques to enhance noise resistance and reduce the compounded effect that is particular existed in intensity function learning. To our knowledge, we are the first to address and offer a solution for both noise-contaminated event and time predictions.
     \item
     We perform experiments on multiple open-source benchmarks containing synthetic noise and deploy our proposed RDHP framework into an OSAHS scenario with real labeling noise from human experts. The results demonstrate the efficacy of \textbf{RDHP} in both real-world medical domains and open-source benchmarks.

 \end{itemize}


\section{Methods}\label{section:method}

\begin{figure*}[h]
    \centering
    \includegraphics[width=\linewidth,scale=1.00]{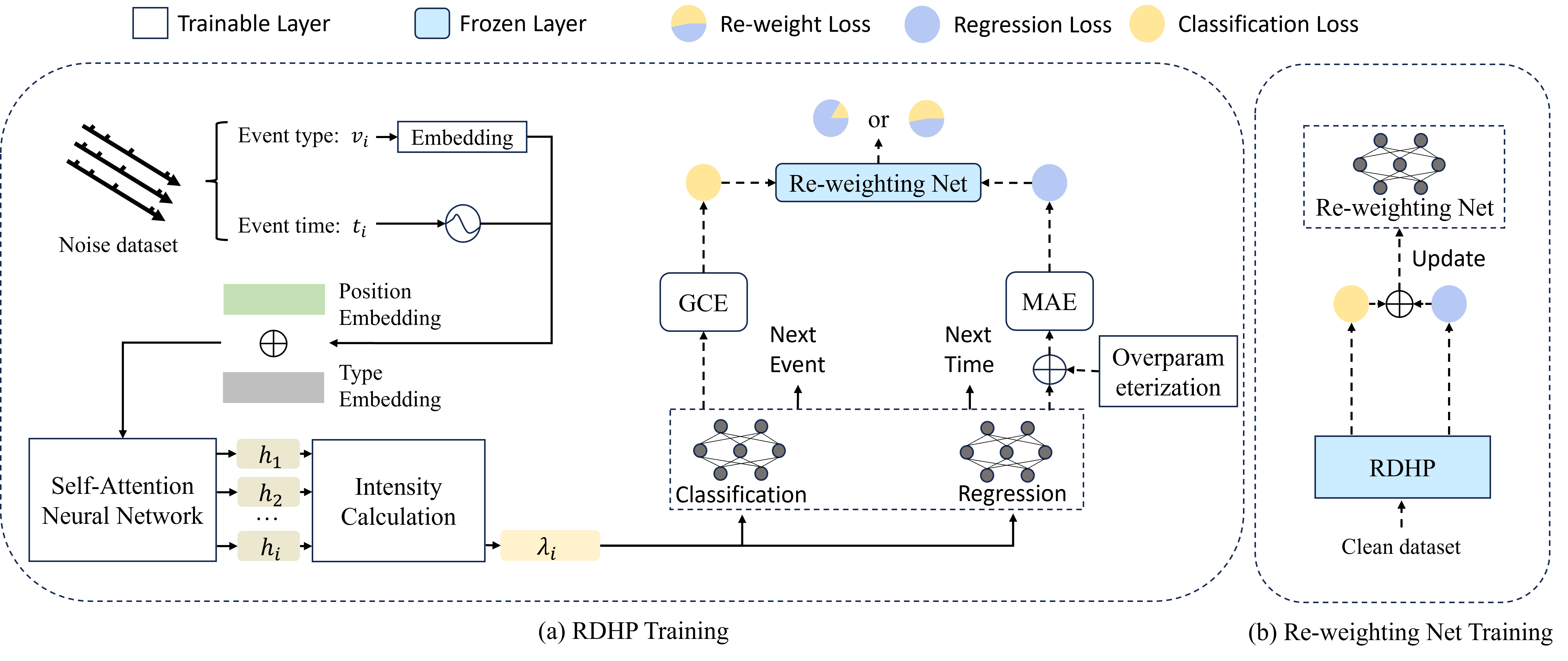}
    \caption{\textbf{(a)} illustrates the training process of RDHP. The re-weighting net is frozen, outputting weights for re-weighting classification and regression losses. \textbf{(b)} depicts the re-weighting net. Before the training of the Hawkes model, the re-weighting net is trained using the losses from a clean dataset. Then, these parameters are frozen for RDHP training.}
    \label{fig:model}
\end{figure*}

\subsection{Preliminary}\label{section:Noise Statement}
We consider a TPP sequence $X = \{ (t_{1},v_{1}), (t_{2},v_{2}), ..., (t_{i},v_{i}) \}$, where each $(t_{i},v_{i})$ denotes the $i$-th event of a time series, with $v_{i}$ representing the event type and $t_{i}$ the arrival time. The goal is to predict the type and occurrence time of the next event based on $X$. In such processes, past events influence future ones, and the occurrence of events is modeled using a conditional intensity function $\lambda ^{\ast } (t)$, which depends on historical events $H_{t}$. This function is defined by the conditional density $f(t\mid H_{t})$ and the cumulative distribution function $F(t\mid H_{t})$ is as follows:
\begin{equation}
\lambda ^{\ast } (t)=\frac{f(t\mid H_{t})}{1-F(t\mid H_{t})}.
\end{equation} Considering a small time interval $[t, t + dt]$, the event probability in this interval is given by:
\begin{equation}
\lambda^{\ast}(t)dt=\mathbb{E}\left[N([t,t+dt])\mid H_{t}\right],
\end{equation}
where $N(\cdot)$ is the counting function for events in the interval. The conditional intensity function can be simulated through the Hawkes process:
\begin{equation}
\lambda ^{\ast}(t) = \mu + \alpha \sum_{t_{i-1}<t}e^{-(t-t_{i-1})},
\end{equation}
where $\mu$ is the base intensity, $\alpha$ is the incentive weight, and each new event increases the conditional intensity, which decays exponentially back to $\mu$ over time. This framework allows for the modeling of event sequences, considering both self-excitement and mutual excitation between different event types.
\begin{equation}
\lambda ^{\ast}(t) = \mu(t) + \alpha \sum_{t_{i-1}<t}\varphi (t-t_{i-1}),
\end{equation} where $\varphi(t) = e^{-(\gamma t)}$ is the density function influenced by historical events, with $\gamma$ controlling the decay rate.

\subsection{The Compounded Effect of Label Noise from Both Events and Occurrence}\label{section:label}

\begin{figure}[h]
    \centering
    \includegraphics[width=0.45\textwidth]{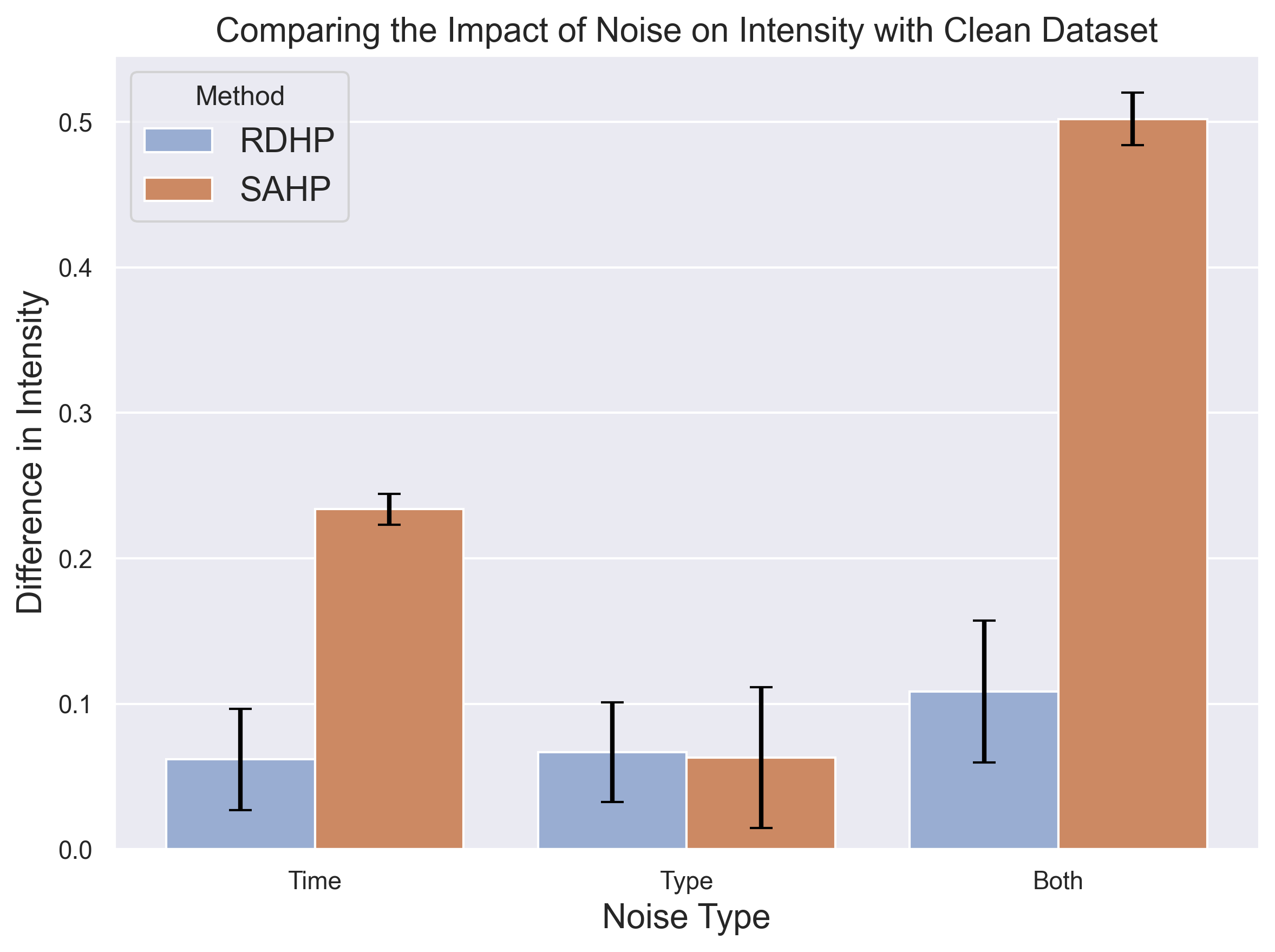}
    \vspace{-5px}
    \caption{Quantitative Impact of Different Noise Types on Deep Hawkes Process Intensity on MIMIC-II dataset. The X-axis represents noise type, while the Y-axis represents the intensity difference between clean and noisy datasets under the same random seed with 30\% uniform noise.}
    \label{fig:intensity_diff}
\end{figure}

\vspace{25px}

In real-world scenarios, however, there is typically some noise in the Hawkes process. For example, for electronic health records (EHRs) \cite{bogner2018human}, there are a number of instances where the wrong disease type or time point is recorded during the manual recording process. As shown in Figure \ref{fig:noise}, such instances can have a significant impact on the Hawkes process prediction. In this study, we conducted an experiment to examine the impact of label noise on the MIMIC dataset using the SAHP algorithm \cite{zhang2020self}. Initially, we trained a model on a clean dataset, using the values of the intensity layers as a benchmark. Subsequently, we introduced two types of noise: Gaussian noise (with a mean of 0 and a standard deviation of 0.8) to corrupt the time labels, and uniform noise for the event labels. These were used to train two separate models to assess how each type of noise affected the learning of the intensity function.

The final step involved combining both types of label noise to train another model. This helped us understand the combined impact of event and time label noise on the intensity layers. We compared the performance of each model by measuring the deviation of their learned intensity layers from those of the clean dataset, as illustrated in Figure \ref{fig:intensity_diff}.

Our findings reveal that both event and time label noise independently affect the learning of intensity layers, leading to errors. More notably, the concurrent presence of both noise types leads to a compounded effect, resulting in errors that exceed the sum of their individual impacts. This complex interaction between event and time label noise hasn't been extensively explored in prior studies. Therefore, developing a robust learning framework that effectively counters the detrimental effects of both types of label noise is crucial.

\subsection{Deep Hawkes process}\label{section:DHP} 

We use a linear embedding layer to encode event types of a time series, transforming one-hot embeddings into a vector \(S_{j}\) using a matrix \(W_{s}\):

\begin{equation}
S_{j} = I_{j} W_{s},
\end{equation}

where \(I_{j}\) is the one-hot embedding of event type \(j\). Event occurrence times are encoded using a sinusoidal function, following the method by \cite{zhang2020self}:

\vspace{15px}

\begin{equation}
T_{j} = \sin(\omega_{k}\times i  + w_{k}\times t_i),
\end{equation}

where \(\omega_{k}\) and \(w_{k}\) represent angular frequency and phase shift, respectively. The final embedding is a combination of these two encodings:

\begin{equation}
\label{eqn:embeding}
x_{i} = S_{j} + T_{j}. 
\end{equation} 

To model event intensity, the model uses self-attention mechanisms, as defined by \cite{vaswani2017attention}:

\begin{equation}
A(Q,K,V) = \text{softmax}\left(\frac{QK^{T}}{\sqrt{d}}\right)V, 
\end{equation}

where \(Q\) and \(K\) are derived from linear transformations of \(x_{i}\), and \(d\) is the dimension of \(K\). The historical information of events is encoded using a hidden vector:

\begin{equation}
\label{eqn:hidden_state}
h(t_{i+1},o) = \frac{\sum_{i=1}^{i}f(x_{i+1}, x_{i})g(x_i)}{\sum_{i=1}^{i}f(x_{i+1}, x_{i})},
\end{equation}

where \(f(\cdot)\) is a Gaussian embedding and \(g(\cdot)\) is a linear transformation. Using this hidden vector, we calculate the conditional intensity function parameters to model Hawkes process:

\begin{equation}
\label{eqn:intensity_func}
\begin{aligned}
\mu_{i+1, o} &= \
{\rm gelu}(h(t_{i+1},o)W_{\mu }), \\
\alpha_{i+1, o} &= {\rm gelu}(h(t_{i+1},o)W_{\alpha }), \\
\gamma_{i+1,o} &= {\rm softplus}(h(t_{i+1},o)W_{\gamma }),
\end{aligned}
\end{equation}

where \({\rm gelu}(\cdot)\) and \({\rm softplus}(\cdot)\) are activation functions. The intensity function \(\lambda_{o}\) reflects the historical event's effect on the next event for \(t \in (t_i, t_{i+1}]\):

\begin{equation}
\lambda_{o} = {\rm softplus}(\mu_{i+1,o} + (\alpha_{i+1,o} - \mu_{i+1,o})e^{-\gamma_{i+1,o}(t-t_{i})}).
\end{equation}

Finally, we use hybrid fully connected MLPs for regression (time prediction) and classification (event-type) based on \(\lambda\). This approach aligns with \cite{zhang2020self} and has shown effectiveness in various real-world TPP scenarios. The hybrid architecture is shown in Figure \ref{fig:model}.

\subsection{Robust Deep Hawkes Process}\label{section:Noise Robust Layer}

The historical event's influence on subsequent events can be analyzed using the deep Hawkes process. This analysis uncovers hidden details that can be instrumental in predicting upcoming events. However, as introduced in Section \ref{section:label}, real-world point process data often comes with inaccuracies regarding the event type and timing, which induces a significant compounded error effect during hidden intensity layer learning. Hence, in this section, we begin by addressing the label noise issue, both in event classification and time regression. We approach this through targeted adjustments in loss functions and network design. Subsequently, we develop a loss balance network, similar to the model discussed in \cite{meta-weight-net}, which is designed to effectively reduce compounded errors by ensuring balanced updates.

\subsubsection{Mitigate Label Noise Effects in Events Learning} \label{section:methods_eve}

In standard deep Hawkes processes, such as SAHP or THP, the frequently employed cross-entropy loss function is not particularly resistant to label noise \cite{zhang2018generalized}. For handling label noise in classification, numerous historical approaches have been proposed for modifying loss functions in the study of label noise. In RDHP, we implement generalized cross-entropy (GCE) loss for event-type prediction \cite{zhang2018generalized}. GCE combines the advantages of MAE noise immunity performance and categorical cross-entropy (CCE) loss fast convergence, and the effectiveness of GCE has a theoretical guarantee from the gradient perspective. To encounter the event type error, we implement the GCE as the loss function of the event type prediction layer:

\begin{equation}
\mathcal{L}^{v}=\frac{1-(M_{e}^{u}(\lambda))^{\beta}}{\beta},
\end{equation}%

where $\beta \in (0,1]$ is a predefined temperature hyperparameter and $M_{e}^{u}(\cdot)$ denotes the $u$-th element of the MLP prediction output, which target is one. $\beta$ controls the degree of GCE loss generalization on CCE and MAE. When $\beta=1$, $\mathcal{L}^{v}$ becomes MAE loss, which is insensitive to outliers, but the gradient decreases too slowly when approaching model convergence, hence extending the model's convergence time \cite{wang2009mean}. When $\beta=0$, $\mathcal{L}^{v}$ approaches CCE loss, which converges quickly but remains unstable for noise data. Training RDHP under GCE loss can exploit the trade-off and enhance noise-resistant performance.

\subsubsection{Mitigate Label Noise Effects in Occurrence Time Learning} \label{section:methods_occ}

For event occurrence time prediction, the noise is imposed on the occurrence time point. We design the timing label noise mitigation part by extending the method proposed by \cite{pmlr-v162-liu22w} in RDHP. The basic idea is to introduce a hyperparameter on each data point to overparameterize the model. Then, during the model learning process, the bias induced by the label noise can be mainly learned by these hyperparameters, and the main model can remain highly generalized. This learning process can significantly reduce the impact of noisy data \cite{pmlr-v162-liu22w}.

Hence, in the timing prediction part of RDHP, our goal is to minimize the difference between $M_{t}(\lambda) +p_{i}$ and $t_{i}$, where $M_{t}(\lambda)$ is the output of the time prediction layer. Based on the sparse overparameterization trick\cite{pmlr-v162-liu22w}, we can formulate the parameter $p_{i}$ by:

\begin{equation}
p_{i}=m_{i}\odot m_{i}\odot t_{i} - n_{i}\odot n_{i}\odot (1-t_{i}),
\end{equation}

where $m_{i} \in [-1,1]$ and $n_{i} \in [-1,1] $ denote two data-specific parameters, and $\odot$ is element-wise production. Please note that since the purpose of the hyperparameter term $p_{i}$ is to learn the noise information during the training process, $p_{i}$ is only optimized in the training process and dropped in inference. To further improve the robustness of $M_t$ in handling large outliers, which might be the data points with large label noise, we implement MAE loss as a time-prediction loss function to optimize this model:

\begin{equation}
\mathcal{L}^{t}=\left |M_{t}(\lambda)+p_{i}-t_{i}\right |.
\end{equation}%

\begin{algorithm}[tb]
    \caption{Robust Deep Hawkes Process}\label{alg:RDHP}
    \textbf{Input}: Noisy Dataset: $D_n$ with $X = \left \{ (t_{1},v_{1}),(t_{2},v_{2}),...,(t_{i},v_{i}) \right \}$; Clean Dataset: $D_c$; Epoch: $N$; Over-parameters: $m$ and $n$; Learning rate $\tau$; Learning rate of over-parameters $\tau_{m}$,$\tau_{n}$; Intensity encoding model $M(\cdot;\theta)$;Re-weighting Net:$ r(\cdot)$; MLP layer for event type: $M_{e}(\cdot;\theta_{e})$; MLP layer for event occurrence time: $M_{t}(\cdot;\theta_{t})$. \\
    \textbf{Output}: Network parameters:$\theta$, $\theta_{u}$, $\theta_{t}$, $m$ and $n$. Next event: $(t_{i+1},v_{i+1})$.
    \begin{algorithmic}[1] 
        \STATE Let Gaussian distribution for $u_{i}$ and $v_{i}$ with zero-mean and s.t.d. $1e-8$.
        \STATE Compute position embedding and event type embedding to obtain $x_{i}$ based on Eqn.~\eqref{eqn:embeding}.
        \FOR{each iteration on minibatch $d_n$ }
        \STATE{}
		\textbf{Hidden representation learning}\\
		\STATE{Calculate the impact of all historical events using the self-attention mechanism to obtain $h(t_{i},v_{i})$ based on Eqn.~\eqref{eqn:hidden_state}}\\
		\textbf{Intensity}
		\STATE{$\lambda _{v}\gets intensity(h(t_{i},v_{i}))$ based on Eqn.~\eqref{eqn:intensity_func}}\\
		\textbf{Inference}
        \STATE{$t_{i+1}\gets M_{t}(\lambda;\theta_{t})$}
        \STATE{$v_{i+1}\gets M_{e}(\lambda;\theta_{e})$}\\
        \textbf{Training the re-weighting net}\\
        \FOR{each iteration on minibatch $d_c$}
            \STATE{Froze $M(\cdot;\theta)$, $M_{e}(\cdot;\theta_{e})$, $M_{t}(\cdot;\theta_{t})$}
            \STATE{Update the Re-weighting Net $r_\epsilon$ with Classification Loss and Regression Loss on clean dataset $d_c$ by Eqn. \eqref{equ:meta_weight_net}\\}
        \ENDFOR\\
        \textbf{1. $M_t$ and $M_e$ update}
        \STATE{$\sigma^v, \sigma^t = r(\mathcal{L}^{v}, \mathcal{L}^{t})$}\\
        \STATE{$\theta_{e}\gets (\theta_{e} - \tau \cdot {\sigma}^{v} \cdot \frac{\partial \mathcal{L}^{v}(M_{e}(\lambda),v_{i+1})}{\partial \theta_{e}})$}
        \STATE{$\theta_{t}\gets (\theta_{t} - \tau \cdot {\sigma}^{t} \cdot \frac{\partial \mathcal{L}^{t}(M_{t}(\lambda)+s_{i},t_{i+1})}{\partial \theta_{t}} )$}\\
        \textbf{2. $s_{i}$ is only updated in training process}\\
		\STATE{$m_{i}\gets P_{[-1,1]}(m_{i} - \tau_{m}\tau \cdot {\sigma}^{v} \cdot \frac{\partial \mathcal{L}^{v}(M_{e}(\lambda),v_{i+1})}{\partial \theta_{e}})$}
        \STATE{$n_{i}\gets P_{[-1,1]}(n_{i} - \tau_{n}\tau \cdot {\sigma}^{t} \cdot \frac{\partial \mathcal{L}^{t}(M_{t}(\lambda)+s_{i},t_{i+1})}{\partial \theta_{t}} )$}\\
        \textbf{3. $M$ update}        \STATE{$\theta\gets (\theta- \tau \frac{\partial \mathcal{L}}{\partial \theta})$}  
		\ENDFOR
    \end{algorithmic}
\end{algorithm}

\subsubsection{Mitigate Compounded Label Noise Effect From Both Events and Occurrence}

As discussed in Section \ref{section:label}, the impact of label noise can be compounded from both events and occurrences, substantially affecting the learning of intensity layers. This, in turn, can lead to significant errors in prediction. Besides, when the dataset is subjected to label noise originating from both events and occurrences, the impact of this noise is also heterogeneous across individual data points. Moreover, different datasets may have different label noise distributions across events and occurrences, which should be addressed adaptively. 

Hence, uniformly applying the label noise mitigation methods introduced in Section \ref{section:methods_eve} on each data might not address such inconsistency. Here, we design a sample re-weighting learning process by assigning varying weights to loss and differentially addressing the varying degrees of noise contamination during the model training process. 

Similar to \cite{meta-weight-net}, we first construct a small clean set $\mathcal{D}_c$ from the whole noise set  $\mathcal{D}_n$ and design re-weighting net
$r$ with parameter $\epsilon$ to generate the sample-wise loss weight $\sigma^t$ for $\mathcal{L}^{t}$  and $\sigma^v$ for $\mathcal{L}^{v}$ :

\begin{equation}
    \sigma^t, \sigma^v = r_\epsilon(\mathcal{L}^{v}, \mathcal{L}^{t}).
\end{equation}

Hence, to optimize the main model $\bar{M}$ with parameter $\theta$ which are introduced in Section \ref{section:DHP} and \ref{section:methods_occ}, we can perform a bi-level optimization approach by minimizing the following two loss on separate dataset:

\begin{equation}
\label{equ:meta_weight_net}
\begin{aligned}
\mathcal{L}^{c}_{\mathcal{D}_c, \epsilon} &= \sigma^v\mathcal{L}^{v}_{\mathcal{D}_c} + \sigma^t\mathcal{L}^{t}_{\mathcal{D}_c},\\
\mathcal{L}^{n}_{\mathcal{D}_n, \theta} &= \sigma^v\mathcal{L}^{v}_{\mathcal{D}_n} + \sigma^t\mathcal{L}^{t}_{\mathcal{D}_n},
\end{aligned}
\end{equation} where the $\epsilon$ is optimized by the $\mathcal{L}^{c}_{\mathcal{D}_c, \epsilon}$ on small clean dataset $\mathcal{D}_c$ and $\theta$ is optimized by the $\mathcal{L}^{n}_{\mathcal{D}_n, \theta}$, iteratively. 

To stabilize the whole training procedure of RDHP, the optimization is carried out in four sequential stages. In the first stage, we focus on updating parameters in $M_t$ and $M_e$. The second stage involves the optimization of the regularization parameters of $m$ and $n$. In the third stage, we update the parameters of $M$. Finally, in the fourth stage, we update the parameters of $r$. Details of this multi-stage optimization process are thoroughly described in Algorithm \ref{alg:RDHP}.


\section{Experimental results}\label{section:experiment}

In this section, the noise resistance of our proposed framework is evaluated using both synthetic and real-world clinical OSAHS datasets provided by the collaborative university hospital. 

\begin{table}
    \caption{Dataset description}
    \vspace{10px}
    \centering
    \begin{tabular}{lrrr}
        \toprule
        Parameters & MIMIC-II & MIMIC-III & OSAHS \\
        \midrule
        Types & 75 & 19 & 4 \\
        Max Length & 706 & 46 & 207\\
        Train & 527 & 1869 & 67\\
        Validation & 58 & 234 & 26\\
        Test & 65 & 234 & 26\\
        \bottomrule
    \end{tabular}
    \label{table:dataset}
\end{table}

\begin{table*}[h]
\caption{$F_{1}$ score and RMSE for the Deep Hawkes Process in the presence of both label noise of event and occurrence time.}
\vspace{10px}
\centering
\resizebox{\textwidth}{!}{
\begin{tabular}{ccccccccccc}
\toprule
\multirow{2}*{Model} & \multicolumn{4}{c|}{Uniform} & \multicolumn{3}{c|}{Flip} & \multicolumn{3}{c}{Flip2} \\  
 &  {0\%} &  {10\%} & {20\%} & \multicolumn{1}{c|}{30\%} &  {10\%} &  {20\%} & \multicolumn{1}{c|}{30\%} &  {10\%} &  {20\%} &  {30\%} \\ \midrule

 & \multicolumn{9}{c}{MIMIC-II $F_{1}$(\%)} \\ \midrule
THP & 40.28 $\pm$ 0.63 & 25.72 $\pm$ 1.79 & 23.83 $\pm$ 0.94 & \multicolumn{1}{c|}{14.59 $\pm$ 1.63} & \textbf{28.97 $\pm$ 0.92} & 21.83 $\pm$ 0.81 & \multicolumn{1}{c|}{17.24 $\pm$ 2.03} & 21.97 $\pm$ 0.42 & 19.81 $\pm$ 0.34 & 17.71 $\pm$ 0.92 \\
SAHP & 36.90 $\pm$ 0.11 & 21.90 $\pm$ 1.94 & 19.06 $\pm$ 2.71 & \multicolumn{1}{c|}{15.59 $\pm$ 3.36} & 20.56 $\pm$ 1.69 & 17.43 $\pm$ 1.82 & \multicolumn{1}{c|}{14.12 $\pm$ 2.35} & 20.91 $\pm$ 1.28 & 16.85 $\pm$ 0.96 & 15.57 $\pm$1.74 \\
RDHP & \textbf{40.41 $\pm$ 3.07} & \textbf{30.96 $\pm$ 1.20} & \textbf{27.24 $\pm$ 1.10} & \multicolumn{1}{c|}{\textbf{20.60 $\pm$ 0.87}} & 26.26 $\pm$ 1.69 & \textbf{23.01 $\pm$ 1.30} & \multicolumn{1}{c|}{\textbf{20.37 $\pm$ 1.71}} & \textbf{26.45 $\pm$ 2.10} & \textbf{23.33 $\pm$ 1.19} & \textbf{19.29 $\pm$ 2.03} 

\\ \midrule

 & \multicolumn{9}{c}{MIMIC-II RMSE} \\ \midrule
THP & 0.82 $\pm$ 0.01 & 1.071 $\pm$ 0.001 & 1.076 $\pm$ 0.007 & \multicolumn{1}{c|}{1.084 $\pm$ 0.003} & 1.067 $\pm$ 0.003 & 1.072 $\pm$ 0.002 & \multicolumn{1}{c|}{1.810 $\pm$ 0.007} & 1.061 $\pm$ 0.004 & 1.083 $\pm$ 0.013 & 1.089 $\pm$ 0.0012 \\
SAHP & 3.89 $\pm$ 2.74 & 9.52 $\pm$ 1.23 & 10.83 $\pm$ 1.63 & \multicolumn{1}{c|}{11.34 $\pm$ 1.38} & 10.05 $\pm$ 0.62 & 11.68 $\pm$ 1.03 & \multicolumn{1}{c|}{12.53 $\pm$ 1.54} & 10.18 $\pm$ 0.25 & 10.66 $\pm$ 0.57 & 11.52 $\pm$ 0.17 \\
RDHP & \textbf{0.60 $\pm$ 0.10} & \textbf{0.76 $\pm$ 0.09} & \textbf{0.88 $\pm$ 0.06} & \multicolumn{1}{c|}{\textbf{0.84 $\pm$ 0.13}} & \textbf{0.82 $\pm$ 0.03} & \textbf{0.88 $\pm$ 0.07} & \multicolumn{1}{c|}{\textbf{1.08 $\pm$ 0.06}} & \textbf{0.89 $\pm$ 0.01} & \textbf{0.93 $\pm$ 0.14} & \textbf{0.92 $\pm$ 0.07} \\ \midrule

 & \multicolumn{9}{c}{MIMIC-III $F_{1}$(\%)} \\ \midrule
THP & 19.73 $\pm$ 1.02 & 14.25 $\pm$ 2.1 & 9.48 $\pm$ 0.93 & \multicolumn{1}{c|}{7.86 $\pm$ 0.74} & 12.57 $\pm$ 0.24 & 11.78 $\pm$ 0.72 & \multicolumn{1}{c|}{10.34 $\pm$ 0.90} & 15.29 $\pm$ 0.43 & 13.95 $\pm$ 0.73 & 12.74  $\pm$ 1.13 \\
SAHP & 30.60 $\pm$ 0.72 & 9.46 $\pm$ 1.93 & 8.11 $\pm$ 1.71 & \multicolumn{1}{c|}{7.76 $\pm$ 1.66} & 8.23 $\pm$ 1.42 & 7.72 $\pm$ 2.99 & \multicolumn{1}{c|}{8.62 $\pm$ 1.51} & 8.59 $\pm$ 1.52 & 7.91 $\pm$ 0.77 & 7.45 $\pm$ 3.21 \\
RDHP & \textbf{34.58 $\pm$ 0.98} & \textbf{26.24 $\pm$ 2.12} & \textbf{24.43 $\pm$ 2.04} & \multicolumn{1}{c|}{\textbf{23.45 $\pm$ 2.67}} & \textbf{30.48 $\pm$ 1.44} & \textbf{23.13 $\pm$ 1.26} & \multicolumn{1}{c|}{\textbf{21.88 $\pm$ 1.96}} & \textbf{27.15 $\pm$ 0.68} & \textbf{21.53 $\pm$ 1.39} & \textbf{19.76 $\pm$ 2.10} \\ \midrule

 & \multicolumn{9}{c}{MIMIC-III RMSE} \\ \midrule
THP & 1.57 $\pm$ 0.002 & 1.58 $\pm$ 0.006 & 1.63 $\pm$ 0.03 & \multicolumn{1}{c|}{1.71 $\pm$ 0.07} & 1.58 $\pm$ 0.01 & 1.577 $\pm$ 0.003 & \multicolumn{1}{c|}{1.575 $\pm$ 0.003} & 1.60 $\pm$ 0.001 & 1.64 $\pm$ 0.031 & 1.58  $\pm$ 0.006\\
SAHP & 1554 $\pm$ 1042 & 2795 $\pm$ 1012 & 2960 $\pm$ 1240 & \multicolumn{1}{c|}{3121 $\pm$ 1140} & 3061 $\pm$ 942 & 3119 $\pm$ 1007 & \multicolumn{1}{c|}{3494 $\pm$ 989} & 2995 $\pm$ 995 & 3386 $\pm$ 972 & 3667 $\pm$ 923 \\
RDHP & \textbf{0.78 $\pm$ 0.05} & \textbf{0.82 $\pm$ 0.04} & \textbf{0.87 $\pm$ 0.04} & \multicolumn{1}{c|}{\textbf{1.00 $\pm$ 0.07}} & \textbf{0.81 $\pm$ 0.09} & \textbf{0.82 $\pm$ 0.03} & \multicolumn{1}{c|}{\textbf{0.93 $\pm$ 0.05}} & \textbf{0.84 $\pm$ 0.02} & \textbf{0.91 $\pm$ 0.04} & \textbf{0.98 $\pm$ 0.03} \\ \bottomrule
\end{tabular}%
}
\label{tab:classification}
\end{table*}

\subsection{Dataset}
\label{sec:dataset}

\textbf{The datasets}: we use two public medical benchmarks to analyze the robustness of RDHP. The datasets we selected are the following, and the details are shown in Table \ref{table:dataset}. The MIMIC-II and MIMIC-III are two widely recognized public datasets that contain extensive records of hospital visits over an extended period. These datasets comprehensively document each visit, capturing key details such as the purpose of the visit and the specific time it occurred. The detail of these datasets is as follows:
\begin{itemize}
    \item 
    \textbf{MIMIC-II}: This dataset is an electronic medical record containing hospitalizations of patients over a seven-year period. Each patient's visit history is considered as a time series, with the events comprising occurrence time points and diagnoses \cite{saeed2002mimic}.     
    \item 
    \textbf{MIMIC-III}: This dataset encompasses information relevant to patients in the intensive care unit (ICU) of a large-scale tertiary hospital \cite{johnson2016mimiciii}. We have selected a dataset from \cite{ru2022sparse} that captures the medical service types received by patients. By combining the patient-received medical service types with their corresponding timestamps, a temporal point process is formulated. This process is employed to predict the subsequent patient visit time and the corresponding medical service type to be administered.
    \item 
    \textbf{OSAHS dataset}: The OSAHS dataset is comprised of OSAHS patients, who sleep at night with different snoring types and may have different noise disturbances in the real environment. We extract the corresponding snoring events and observe a sparse point distribution of snoring events throughout a single night of patient sleep, which is considered as a TPP. Each patient's snoring is recorded for a complete night of 8 hours, and the entire night of sleep is divided into four parts, with each sequence length of 2 hours. Snoring event types are classified as central sleep apnea (CSA), mixed sleep apnea syndrome (MSAS), obstructive apnea (OA), and hypopnea. The last event point in the sequence is used as the target to predict the event type and event occurrence time of the next snoring event. The discrimination of snoring events relies on the subjective judgment and analysis of medical professionals and is influenced by the performance of PSG instruments and environmental factors during data collection. In the OSAHS dataset presented in Table \ref{table:dataset}, the longest sequence spans two hours and encompasses a total of 207 recorded events. Extrapolating to a full 8-hour night of sleep results in a total of 1656 events. If the misclassification rate of snoring events exceeds $3\%$, it can potentially lead to the erroneous diagnosis of mild OSAHS in patients.
\end{itemize}

\begin{figure}[h]
    \centering
\includegraphics[width=0.47\textwidth]{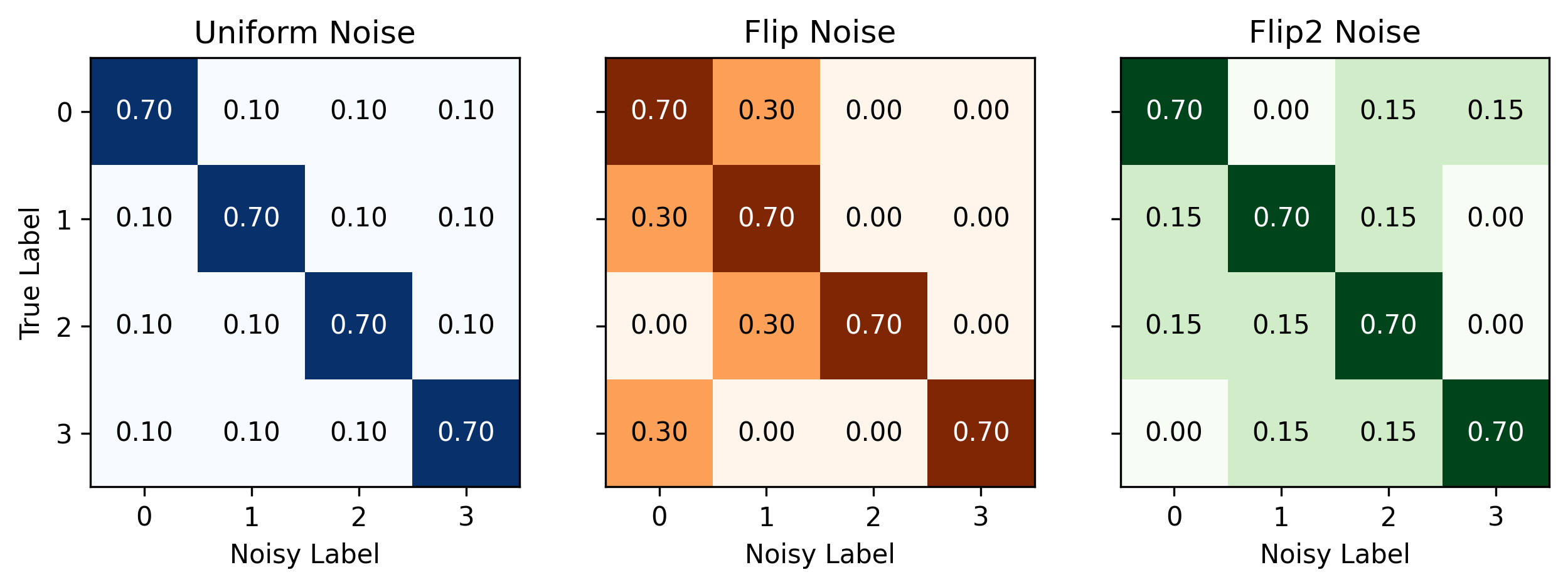}
    \caption{Label Noise Types Applied in the Experiment ($K=4$, $p=0.3$).}
    \label{fig:noise_matrix}
\end{figure}

\subsection{Noise setup}\label{section:noise set}

To assess the model's robustness to label noise, we introduce noise into the training set, while maintaining the integrity of the validation and test sets. The dataset is modified with synthetic noise using a noise probability \( p \). Consider a time series \( X = \{ (t_{1},v_{1}), (t_{2},v_{2}), ..., (t_{i},v_{i}) \} \), where \( v_{i} \) represents the event type of the \( i \)-th event, taking values from \( 1 \) to \( K \), and \( t_{i} \) signifies the event occurrence time within the range \( [0,t_{max}] \). Label noise means that for each event \( (t_{i},v_{i}) \), there is a probability \( p \) that either \( t_{i} \) or \( v_{i} \), or both, may be incorrect. With a noise probability of \( p \), the timestamp \( t_{i} \) could be modified to \( t_{i} + t_{noise} \), where \( t_{noise} \) is a Gaussian-distributed variable with a mean of \( 0 \) and a standard deviation of \( 0.8 \). This alteration can lead to changes in the chronological order of events, as illustrated in Figure. \ref{fig:noise}.

We incorporate three distinct forms of label noise into our analysis: Uniform noise, Flip noise, and Flip2 noise. The matrices illustrating how each type of noise affects the labels are depicted in Figure \ref{fig:noise_matrix}. For a more comprehensive explanation of these noise types, including their specific characteristics and how they are applied to the data, please refer to Appendix \ref{Ap:noise_set}.

\subsection{Experiment Setup and Implementation}

Before introducing noise, in a synthetic experiment, we randomly extract 5\% clean datasets from the training set, which are used for optimizing the re-weighting net. For different noise dataset scenarios, we simulated synthetic datasets with a noise rate of $p$, and the noise was added by introducing Gaussian distribution on timestamps and the three aforementioned types of label noise. This corresponds to the situation where experts incorrectly record time points and event types during data labeling.

We utilized a dataset of OSAHS patients obtained in a noisy environment and manually labeled by human experts. This dataset is used to test the robustness of RDHP under TPP with real human label noise. The dataset is divided into $60\%$, $20\%$, $10\%$, and $10\%$ for train, validation, test, and clean, respectively. Here, we invite the human expert to relabel the validation and test sets carefully, which are considered clean sets. For synthetic noise datasets, we divide the dataset into 80$\%$ for training, 10$\%$ for validation, and 10$\%$ for testing.

 \begin{table}[]
    \caption{Experiment results on OSAHS dataset.}
    \begin{center}
        \begin{tabular}{cccc}
        \toprule
        Model  & THP & SAHP & RDHP\\
        \midrule 
        $F_1$(\%)  & 18.04 $\pm$ 7.40& 40.90 $\pm$ 10.77& \textbf{69.41 $\pm$ 8.22}\\
        RMSE  & 0.69 $\pm$ 0.01 & 28.95 $\pm$ 4.24 & \textbf{0.59 $\pm$ 0.07}\\
        \bottomrule
        \end{tabular}
        \label{table:realdata}
    \end{center}
 \end{table}
 
Since the event type is a multi-classification problem, as shown in Table \ref{table:dataset}, we choose Macro $F_1$ score to evaluate the noise robustness of RDHP. For the event occurrence time prediction, we choose RMSE to evaluate the difference between the predicted and true values.

We compare RDHP with two state-of-the-art deep Hawkes process models, which are self-attentive Hawkes process (SAHP) \cite{zhang2020self} and transformer Hawkes process (THP) \cite{zuo2020transformer}. The noise robustness of the model is measured by comparing the model's Macro $F_1$ score for event type and RMSE for event occurrence time regression on a clean test set. All experiments were conducted on the Nvidia RTX 3080 GPU, using the Python 3.10 platform, and trained based on the PyTorch 1.11.0 framework.

\subsection{Main Results}
\label{sect:main_result}

These experiments involved the application of three different noise types and were performed in five separate trials. The average number of iterations for the experiments was 200, with an average runtime of approximately 3000 seconds. The experimental results represent the mean values obtained from five independent trials with random seeds, which have been summarized and listed in Table \ref{tab:classification}. The hyperparameter settings for the RDHP model experiment can be found in Appendix \ref{section:hyperparameter}. 

RDHP achieves the most stable and best classification $F_1$ on MIMIC-II in all noise cases. It converges to the best RMSE for all the event occurrence time prediction. It also stabilizes the training on MIMIC-III and achieves superior noise robustness compared to baseline methods.
   \begin{table}[]
    \caption{Ablation study on the impact of noise. The experiment employed synthetic noise on the MIMIC-II dataset.}
    \begin{center}
    \resizebox{0.47\textwidth}{!}{
        \begin{tabular}{cccccc}
        \toprule
        \multirow{2}{*}{Dataset}
        &
        \multirow{2}{*}{Noise}
        &
        \multicolumn{2}{c}{SAHP}
        &
        \multicolumn{2}{c}{RDHP} \\
         &   & $F_{1}$(\%) & RMSE & $F_{1}$(\%) & RMSE\\
        \midrule 
        MIMIC-II & 0\% & 36.90 $\pm$ 0.11 & 3.89 $\pm$ 2.74 & \textbf{40.41 $\pm$ 3.07} & \textbf{0.60 $\pm$ 0.10}\\
        \midrule 
        \multirow{2}{*}{Only Time Noise}  & 10\% & 26.76 $\pm$ 0.29 & 10.62 $\pm$ 0.44 & \textbf{39.49 $\pm$ 1.70} & \textbf{0.74 $\pm$ 0.05}\\
         & 20\%  & 27.78 $\pm$ 0.63 & 11.37 $\pm$ 0.73 & \textbf{35.36 $\pm$ 1.04} & \textbf{0.82 $\pm$ 0.08}\\
        \midrule 
        \multirow{2}{*}{Only Label Noise}  & 10\% & 22.42 $\pm$ 1.51 & 8.58 $\pm$ 0.47 & \textbf{32.68 $\pm$ 1.66} & \textbf{0.70 $\pm$ 0.02}\\
         & 20\%  & 17.95 $\pm$ 1.07 & 9.77 $\pm$ 0.62 & \textbf{29.36 $\pm$ 0.84} & \textbf{0.75 $\pm$ 0.04}\\
        \midrule 
        \multirow{2}{*}{Time and Label Noise}  & 10\% & 21.90 $\pm$ 1.94 & 9.52 $\pm$ 1.23 & \textbf{30.96 $\pm$ 1.20} & \textbf{0.76 $\pm$ 0.09}\\
         & 20\%  & 19.06 $\pm$ 2.71 & 10.83 $\pm$ 1.63 & \textbf{27.24 $\pm$ 1.10} & \textbf{0.88 $\pm$ 0.06}\\        
        \bottomrule
        \end{tabular}}
        \label{table:ablation-noise}
    \end{center}
\end{table}
In the uniform noise case, the RDHP model generally exhibits greater robustness in classification compared to THP and SAHP, with a significantly lower RMSE. In summary, the classification performance of SAHP and THP deteriorates under moderate to high noise conditions, resulting in a rapid increase in RMSE. Furthermore, noise has a catastrophic impact on SAHP and THP, particularly in high-noise scenarios, where their classification performance notably declines. This supports our previous assertion that existing deep Hawkes processes may lack robustness in the presence of label noise. However, through robust structural modifications and overparameterization regression, RDHP maintains high accuracy and low RMSE even in high-noise environments.

Table \ref{table:realdata} shows the performance of RDHP with baseline methods on the OSAHS dataset. This dataset is a time series of snoring events taken from OSAHS patients throughout the night, and the presence of marker noise is assumed. After the experiments, we can observe that RDHP achieves superior performance in the real case, on both classification and regression. THP overfits the noise data and demonstrates its unstable performance in the real circumstances under label noise. The classification effect of SAHP in the real environment is also greatly reduced, and the prediction error for the time of event occurrence is large. The results indicate that, so far, deep Hawkes models remain unstable under real-world label noise.

\section{Ablation Study}\label{section:ablation}

In this section, we conduct an ablation study to evaluate the contributions of each component in our RDHP model, particularly focusing on its label noise resilience capability. For event type prediction, the model employs a training loss based on the Generalized Cross Entropy (GCE) and utilizes a dedicated event prediction subnet, denoted as $M_e$. In contrast, time regression is handled by a separate subnet, $M_t$, which applies sparse overparameterization regression. Additionally, we incorporate a re-weighting network, $r$, to ensure balanced training between classification and regression tasks. 

The ablation study is structured to assess the individual impact of these components. Initially, we explored the model's performance under various noise conditions, including scenarios with only temporal noise, only label noise, and situations where both occur simultaneously. This investigation aimed to understand how different types of noise affect the model's robustness. We also examined the convergence patterns of the weights for classification and regression losses during balanced training. This was crucial to ensuring that both aspects of the model were trained effectively. Lastly, we analyzed the role of overparameterization regularization in enhancing the RDHP model’s performance. 

\vspace{10px}

 \begin{figure}[h]
    \centering
    \includegraphics[width=0.45\textwidth]{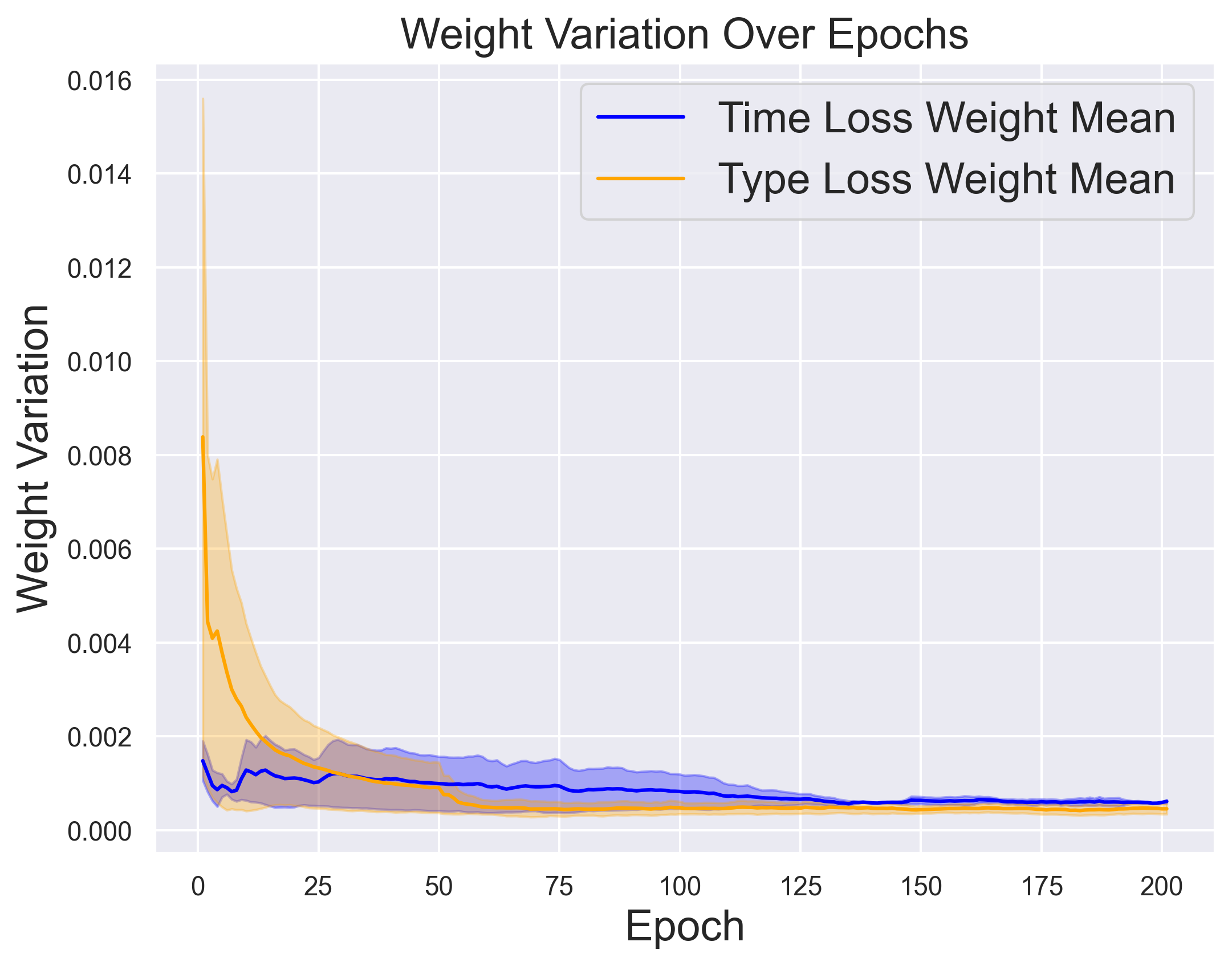}
    \caption{Weight variation under 30\% uniform noise experiment on MIMIC-II dataset.}
    \label{fig:ab_weight}
\end{figure}

\vspace{10px}

\subsection{Impact of Noise}

The experimental results provide insights into the impact of noise on the SAHP and RDHP using different noise scenarios applied to the MIMIC-II dataset.The time point noise is configured following Section \ref{section:noise set}, with a Gaussian distribution for time point noise and uniform noise for label noise. The results are presented in Table \ref{table:ablation-noise}. In the study, the introduction of only occurrence noise significantly affected the RMSE, with a minor reduction in the $F_1$ score. Conversely, when only label noise was introduced, it notably impacted the accuracy of event type prediction, leading to a substantial increase in RMSE as the noise level increased. Moreover, when both time and label noise were present, the performance in terms of $F_1$ score for classification and the RMSE for regression was the poorest. This decline in performance is attributed to the nature of the Hawkes process, a self-exciting model, where noise considerably affects the computation of intensity, as evident in Figure \ref{fig:intensity_diff}. This, in turn, adversely affects the final model's performance in both classification and regression tasks. However, our proposed RDHP demonstrates superior classification performance and maintains a low RMSE under these conditions, indicating its resilience to noise. These findings highlight the Deep Hawkes Process's vulnerability to noise and underline the necessity of implementing noise reduction strategies or robust training methods to ensure dependable performance in practical applications.

\subsection{The Learning of loss weight $\sigma$}

In Figure \ref{fig:ab_weight}, we present the dynamic adjustments of the loss weight $\sigma$ in the RDHP model, considering both time loss and event loss, with 30\% uniform noise. The X-axis represents the progression of training epochs, while the Y-axis tracks the difference $\Delta\sigma$ in weight values between epochs. This visualization reveals a notable pattern in the training phase: the weights assigned to each loss function adaptively adjust over time, eventually reaching a state of equilibrium. The re-weighting network plays a crucial role in this process, effectively managing the balance and mitigating the compounded bias from the learning loss for events and timing. This equilibrium not only enhances the model's performance but also incorporates a systematic approach to mitigating label noise in both the event and timing aspects of the training data. In Table \ref{table:ab-all}, the RDHP's performance in both classification and regression tasks decreases without the re-weighting net.

\subsection{Regularization}

As introduced above, robust regularization utilizes new hyperparameters on each data to separate the noisy information. The results demonstrate that the robustness of the classification task has been slightly enhanced, whereas the regression task has been significantly enhanced. In Table \ref{table:ab-all}, we present a comparative analysis of various models. It is evident that the incorporation of robust regularization leads to a notable reduction in the RMSE. Given that each sequence in the MIMIC-II dataset represents a patient's medical history over a period of 7 years, normalized within the range $[0,5.3846]$, even minor inaccuracies in time prediction can result in significant deviations. This context underscores the exceptional performance of the RDHP model, which maintains an RMSE of approximately 1.0, indicating its high accuracy. Furthermore, the model demonstrates resilience against label noise, showing minimal impact on its regression capability.

\vspace{10px}

\begin{table}[h]
\caption{RMSE and $F_1$(\%) for ablation study results.}
    \begin{center}
    \resizebox{0.47\textwidth}{!}{
        \begin{tabular}{cccc}
        \toprule
        Model & Noise Rate & RMSE & $F_{1}$(\%)\\
        \toprule
        \multirow{2}{*}{SAHP} & 30\% &  11.34 $\pm$ 1.38 & 15.59 $\pm$ 3.36 \\
         & 50\% & 12.12 $\pm$ 0.67 & 12.16 $\pm$ 2.54 \\
        \toprule
        RDHP & 30\% & 1.07 $\pm$ 0.06 & \textbf{21.08 $\pm$ 0.61}\\
        without Regularization & 50\% & 1.27 $\pm$ 0.28 & 16.94 $\pm$ 1.67\\
       \toprule
        RDHP & 30\% & 0.97 $\pm$ 0.15 & 19.14 $\pm$ 0.95\\
        without Re-weighting Net & 50\% & 1.13 $\pm$ 0.19 & 18.61 $\pm$ 1.18\\
       \toprule
       \multirow{2}{*}{RDHP} & 30\% & \textbf{0.84 $\pm$ 0.13} & 20.60 $\pm$ 0.87\\
        & 50\% & \textbf{1.03 $\pm$ 0.16} & \textbf{18.43 $\pm$ 1.51} \\
       \toprule
        \end{tabular}}
        \label{table:ab-all}
    \end{center}
 \end{table}

\section{Conclusion}\label{section:conclusion}

In this study, we introduce a novel robust deep Hawkes process (RDHP) model specifically designed to handle corrupted training labels. Prior research has largely overlooked the compounded bias arising from label noise in both event types and their occurrence times. To the best of our knowledge, our RDHP model is the first to tackle this challenge. It improves the resilience of event classification and time prediction in noisy environments through a robust loss function, sparse overparameterization regularization, and loss re-weighting strategies. We demonstrate the limitations of existing Hawkes process models for handling label noise through experiments on synthetic and real OSAHS datasets. Our RDHP model consistently outperforms traditional models, achieving higher classification accuracy and a lower RMSE. Further, we conduct ablation studies to assess the impact of each component of our model. This research paves the way for enhancing robustness in broader point process models, offering a significant advancement in handling real-world data with inherent label noise. Future works will focus on further refining accuracy in noisy conditions and expanding applicability to a wider range of practical noisy scenarios.


\begin{ack}

This work was supported by Shanghai Municipal Natural Science Foundation (23ZR1425400), the National Natural Science Foundation of China (62102241), Eye \& ENT Hospital's double priority project A (YGJC026 to Dr Wei and Dr Huang).

\end{ack}






\bibliography{Main}

\clearpage

\appendix
\onecolumn

\section{Related Work}\label{section:related work}
\subsection{Temporal Point Process}\label{section:TPP}

The temporal point process (TPP)\cite{gonzalez2016spatio,fuster1995temporal,hawkes1971spectra} is an approach of modeling time series that employs point patterns, which can be employed to model historical events. It is assumed that events in the point model occur instantaneously and can be represented as points on a time series. These points are interconnected, and what occurred in the past may have an impact on what will occur in the future \cite{rasmussen2018lecture}.

The conventional point process methods include the poisson process \cite{lawless1987regression}, the self-correcting process \cite{self-correcting}, and the Hawkes process \cite{hawkes1971spectra}. The Hawkes process is a self-exciting model, which means that previous events have an incentive effect on future events, and this incentive effect decays with time, in accordance with the real world. Hawkes process is commonly used in earthquake prediction, finance, and disease prediction \cite{hawkes1974cluster}.

With the recent advances in deep learning technology in recent years, the deep Hawkes process technique has attracted considerable interest. \cite{zhang2020self} proposes the SAHP model by combining the Hawkes process with the attention mechanism to increase the connection between event points in a sequence. \cite{zuo2020transformer} combines the Hawkes process with Transformer to capture long-term dependencies using a self-attention mechanism while increasing computational efficiency. However, the aforementioned models' assumptions are modeled in a noise-free situation, i.e., the event type and event time point are precisely recorded. Real-world data is always contaminated with various types of labeling noise, such as human labeling errors and technical recording techniques, significantly affecting the learning process's efficiency and stability.\cite{frenay2013classification,lukasik2020does,song2022labelnoise}. 

\subsection{Label Noise}\label{section:label noise}
When considering a time series with noise, there are typically two types of noise: one is caused by the time shift of the event time point (i.e., label noise of occurrence time), and the other is the event type being recorded incorrectly (i.e., label noise of the event). Both of these types of noise are prevalent in real-world scenarios and might have an impact on the robustness of the model.

A number of methods have been developed to improve the model's robustness under label noise of event. Modifying the model's structure, adjusting sample selection, incorporating regularization, and designing loss functions are typical techniques for improving model robustness \cite{song2022labelnoise}.

\textit{Noise modeling} \cite{xiao2015noiselayer} proposes a method employing two independent networks, both for predicting noise type and label shift probability, respectively. After the training step, both networks are trained with numerous noisy labels and a small amount of clean data, making their adaptability to more general architectures challenging.

\textit{Sample selection} the Co-teaching family \cite{han2018coteaching} \cite{yu2019coteaching+} \cite{wei2020jocor} as the main representative, multiple network structures are used to filter each other, and there is the problem of increasing the training parameters and decreasing the training efficiency. 

\textit{Robust regularization} data augmentation \cite{dataaugment}, weight decay \cite{weightdecay}, dropout \cite{srivastava2014dropout}, and batch normalization \cite{ioffe2015batchnormalization}, can improve model robustness, but these methods suffer from reduced generalization when the noise is high \cite{tanno2019learning}. The loss function is reviewed in the following. Moreover, the mentioned methods also have a uniform problem, that is, only the label noise of the event is considered without considering the label noise of the occurrence time point.

Recent studies have also extensively focused on the label noise of the event's occurrence time point. \cite{trouleau2019-syn-noise} claim that each sequence has a fixed time shift (synchronous noise), and the causal relationship between the sequences can be reconstructed by modifying the excitation kernel function and estimation method. Subsequently, \cite{trouleau2021cumulants-noise} amend the noise assumption and correct the sequence causality by the cumulative amount of MHP, with each sequence noise belonging to different independent distributions. However, the above-mentioned models have certain limitations: they concentrate solely on the relationship between time sequences without accounting for the simultaneous presence of label noise existing in both event types and occurrence time points, which will have a negative effect on the classification accuracy of the following event.

Categorical cross-entropy loss (CCE) is the most commonly used loss function in event classification tasks due to its fast convergence and high generalization capabilities. \cite{ghosh2017MAEloss} demonstrate that in noisy tasks, MAE has superior generalization performance than CCE and is more robust under noise by creating a risk minimization model. However, if the data is more complicated and the event types are more diverse, MAE is difficult to converge, and its generalization performance will subsequently decrease. Subsequently, \cite{zhang2018generalized} derives the GCE loss function from the gradient properties of MAE and CCE. It has both the fast convergence property of CCE and the robustness of MAE in terms of noise performance.

We present a deep robust Hawkes process (RDHP) that combines a robust loss function and robust regularization in order to deal with the simultaneous presence of label noise in both event types and occurrence time points. The model encodes a sequence of event points with event type and occurrence time noise, produces point-to-point attention scores, and then models them using the Hawkes process in order to obtain historical event hidden information. GCE is presented as a loss function for event prediction. The prediction of occurrence time points, which is more probable to be a regression problem, employs the more robust MAE loss function. Meanwhile, robust regularization is implemented, and new hyperparameters are added to create a data model for learning clean data by separating noise information. The overview of our proposed model is shown in Figure \ref{fig:model}.

\section{Noise Setting}\label{Ap:noise_set}
Here we describe three types of label noise. Consider a time series $X =  { (t_{1},v_{1}), (t_{2},v_{2}), ..., (t_{i},v_{i}) }$, where $v_{i}\in { 1,2,3,...,K }$ denotes the event type of the $i$-th event, and $t_{i}\in \left [0,t_{max}\right ]$ represents the event occurrence time.

\begin{itemize}
    \item 
    \textbf{Uniform Noise}: Uniform noise refers to each event type having an equal probability of $p$ to transform into any of the other $K-1$ classes. Specifically, the dataset with a corruption probability of $p$ means that $v_{i}$ has probability $p$ being corrupted into other $K-1$ labels in event point $(t_{i},v_{i})$, which is a one-to-$K-1$ type of label noise. 
    \item 
    \textbf{Flip Noise}: Flip noise involves pairing event types one by one and corrupting them with a probability of $p$, causing a specific class to be fixedly transformed into another. Specifically, the dataset with a corruption probability of $p$ means that $v_i$ is flipped to another label, $u_i \ne v_{i}$. Therefore, $v_i$ has a probability of $(1-p)$ to remain the $v_i$, which is a one-to-one type of label noise. 
    \item 
    \textbf{Flip2 Noise}: Flip2 noise refers to the random corruption of the true label into two other classes with a probability $p$. Specifically, the dataset with a corruption probability of $p$ means $v_i$ has probability $(1-p)$ of being the $v_i$ and probability $p/2$ of being corrupted into other two labels $u_i^1 \ne v_{i}$ or $u_i^2 \ne v_{i}$, which is a one-to-two type of label noise.
\end{itemize}

\section{Hyperparameter}\label{section:hyperparameter}
Here we show the hyperparameter we used in the experiments.

\begin{table}[h]
\caption{Hyperparameters used for training each dataset.}
    \begin{center}
    \small
        \begin{tabular}{ccccccccc}
        \toprule
        Dataset  & batch size & learning rate & attention head & attention layer & dropout rate & hidden size & MLP layer & $\beta$ in GCE\\
        \toprule
        MIMIC-II  & 16 & 1e-3 & 8 & 4 & 0.2 & 32 & 3 & 0.7\\
        MIMIC-III  & 32 & 1e-4 & 8 & 4 & 0.2 & 32 & 3 & 0.7\\
        OSAHS  & 8 & 1e-4 & 6 & 3 & 0.1 & 16 & 3 & 0.7\\
        \toprule
        \end{tabular}
        \label{table:hyperparameter}
    \end{center}
\end{table}

\section{Re-weight Net Algorithm}

We show the algorithm of Re-weight Net in Algorithm \ref{alg:re_weight}.

\begin{algorithm}[tb]
    \caption{Training the re-weighting net}\label{alg:re_weight}
    \textbf{Input}: Clean Dataset: $D_c$; Learning rate $\tau$; Intensity encoding model $M(\cdot;\theta)$;Re-weighting Net:$ r(\cdot)$; MLP layer for event type: $M_{e}(\cdot;\theta_{e})$; MLP layer for event occurrence time: $M_{t}(\cdot;\theta_{t})$. \\
    \textbf{Output}: Re-weighting Net:$ r(\cdot)$ with Updated.
    \begin{algorithmic}[1] 
        \FOR{each iteration}
        \STATE{Froze $M(\cdot;\theta)$, $M_{e}(\cdot;\theta_{e})$, $M_{t}(\cdot;\theta_{t})$}\\
        \textbf{Training the re-weighting net}\\
        \STATE{$d_{c}$ $\leftarrow$ SampleMiniBatch($D_c$)}\\
        \STATE{Calculate Classification Loss and Regression Loss on clean dataset $d_{c}$ by Eqn. \eqref{equ:meta_weight_net}\\}
        \STATE{Update the Re-weighting Net $r_\epsilon$\\}
		\ENDFOR
    \end{algorithmic}
\end{algorithm}

\end{document}